\documentclass{ieeeaccess}
\usepackage{amsmath,amssymb,amsfonts}
\usepackage{natbib}
\usepackage{bm}
\usepackage{mathtools}
\usepackage{fixmath}
\usepackage{diagbox}
\usepackage{booktabs}
\usepackage{caption}
\def\BibTeX{{\rm B\kern-.05em{\sc i\kern-.025em b}\kern-.08em
    T\kern-.1667em\lower.7ex\hbox{E}\kern-.125emX}}
\usepackage{ifthen}
\newboolean{showcomments}
\setboolean{showcomments}{true} 
\ifthenelse{\boolean{showcomments}}{
\newcommand{\mynote}[3]{\noindent{\color{#3}\textbf{#1:\xspace} #2}}}
{ \newcommand{\mynote}[3]{}
}
\definecolor{mypurple}{rgb}{0.6,0.4,0.8}

\begin{document}
\history{}
\doi{}

\title{Reinforcement Learning to Optimize the Logistics Distribution Routes of Unmanned Aerial Vehicle}
\author{\uppercase{Linfei Feng}\authorrefmark{1}}

\address[1]{Department of Mathematics, University of Southampton, Southampton, SO17 1BJ, UK (e-mail: lf1a18@soton.ac.uk)}

\markboth
{Author \headeretal: Preparation of Papers for IEEE TRANSACTIONS and JOURNALS}
{Author \headeretal: Preparation of Papers for IEEE TRANSACTIONS and JOURNALS}

\begin{abstract}
Path planning methods for the unmanned aerial vehicle (UAV) in goods delivery has draw great attentions from industry and academics because its flexibility which is suitable for many situations in the "Last Kilometer" between customer and delivery nodes. However, the complicated situation is still a problem for traditional combinatorial optimization methods. Based on the state-of-the-art Reinforcement Learning (RL), this paper proposed an improved method to achieve path planning for UAVs in complex surroundings: multiple no-fly zones. The improved approach leverages the attention mechanism and includes embedding mechanism as the encoder and three different widths of beam search (i.e.,~1, 5, and 10) as the decoders. Policy gradients is utilized to train the RL model for obtaining the optimal strategies during inference. The results show the feasibility and efficiency of the model applying in this kind of complicated situation. Comparing the model with the results obtained by the optimization solver OR-tools, it improves the reliability of the distribution system and has a guiding significance for the broad application of UAVs.
\end{abstract}


\titlepgskip=-15pt

\maketitle
\section{Introduction}
\label{sec:introduction}

\begin{figure}[!htb]
    \centering
    \includegraphics[width=7.5cm]{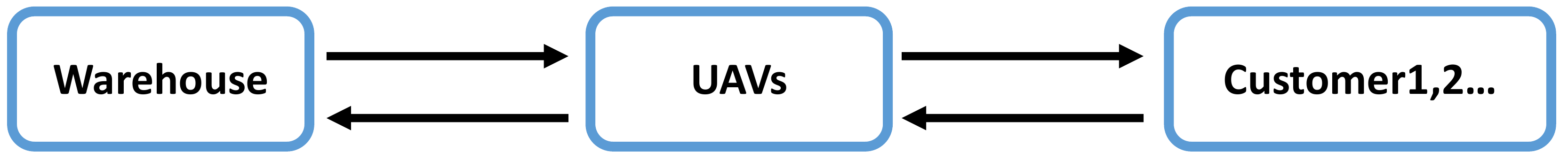}
    \caption{UAV operation process}
    \label{fig:UAV_operation_process}
\end{figure}

Unmanned aerial vehicle has been try to applied in goods delivery in many logistics enterprises, but it not been popularized at present, and still mainly based on experiment and exploration. UAVs logistics distribution refers to the ways of delivery goods through drones in distribution activities. In this process, it replaces other transport equipment to delivering items to customers location and generally small. This research mainly uses such drones as research objects. The specific operation process shown in Figure~\ref{fig:UAV_operation_process}. The drone loads the goods in the warehouse and accepts transportation data transmission. After loading and receiving the data, it enters the self-controlled flight mode. Then it arrives at the destination and landing and unloading the package according to the present flight path. Finally, automatically return to the warehouse. There are various advantages on the UAVs distribution, such as decreasing the labour cost, reducing the pressure on the current congested traffic, quickly respond to the needs of consumers, etc~\citep{zhang2012application}.
Currently, consumers are increasingly focusing on the speed of distribution, especially for the critical file and take-out delivery. In this situation, the use of UAV for distribution has broad application prospects. Thus, many attempts and explorations have been carried out by some companies in this field. For example, Amazon released a new version of the delivery drone 'prime air' this year and planned to send the parcel to buyers through drones within 30 minutes \citep{coombs2019amazon}. An American enterprise 'Wing' began offering deliverable packages to residents of the Christiansburg area of Virginia this year and promise to deliver the goods within a few minutes \citep{WIGGERS2019Wing}.
But on the other hand, drones also have some drawbacks. Drones cannot fly in sensitive areas. Because the drone is controlled by radio, it will have a high impact on other aircraft. We can learn that flight delays caused by drones often occur these years. Therefore, in many countries and regions, no-fly zones have been set up for drones. Besides, since the drone is easily affected by the weather, the flight of the drone has specific safety hazards under severe weather conditions \citep{yu2018unmanned}. Therefore, when encountering such an area, the drone needs to detour or stop flying.

This paper mainly studies the route planning problem of the UAVs distribution. The application scenario is the goods delivery from the distribution centre to the customer, and we do a flight path planning for drones that need to serve multiple customer positions. Considering the UAV has some disadvantages, such as a small load and no-fly zone in some city areas. We set the goods being delivered cannot exceed the UAVs maximum capacity and the flight area also has some constraints. The purpose is to found a method that can plan the shortest flight distance for the UAVs in a situation that meets the characteristics of actual needs.

In the field of artificial intelligence, machine learning methods have shown excellent function in solving a complex problem. But there is not much research on the path planning using this method. In the review of path planning for vehicle and drones, it is found that most of the path planning problems are solved by heuristic algorithms. While the path planning problem is a classic NP problem, although researchers have proposed many ways to solve it, the solution can still be improved. So we plan to use reinforcement learning to solve this problem. On the other hand, various kinds of limitations in the application of UAV distribution are fully considered in this research, which makes the operation of UAV closer to the reality and provides a reference for the development and use of UAV distribution in the future.

The structure of this paper is as follows. The second sections carried out a literature review. It includes the summarizes of the methods of path planning for vehicle and drones that used in different kinds of situations, the introduces of the basic knowledge of reinforcement learning, and the current research status of artificial intelligence applied to path planning. The third section of the paper describes the improved model of path planning, which is established under the framework of reinforcement learning. In the next section, we trained and verified the effectiveness of the model with simulated data that is consistent with current human active characteristics. Finally, the experimental results are analyzed and discussed.

\section{Literature review}
\subsection{Route planning problem}
\begin{figure}[!htb]
  \centering
    \includegraphics[width=6cm]{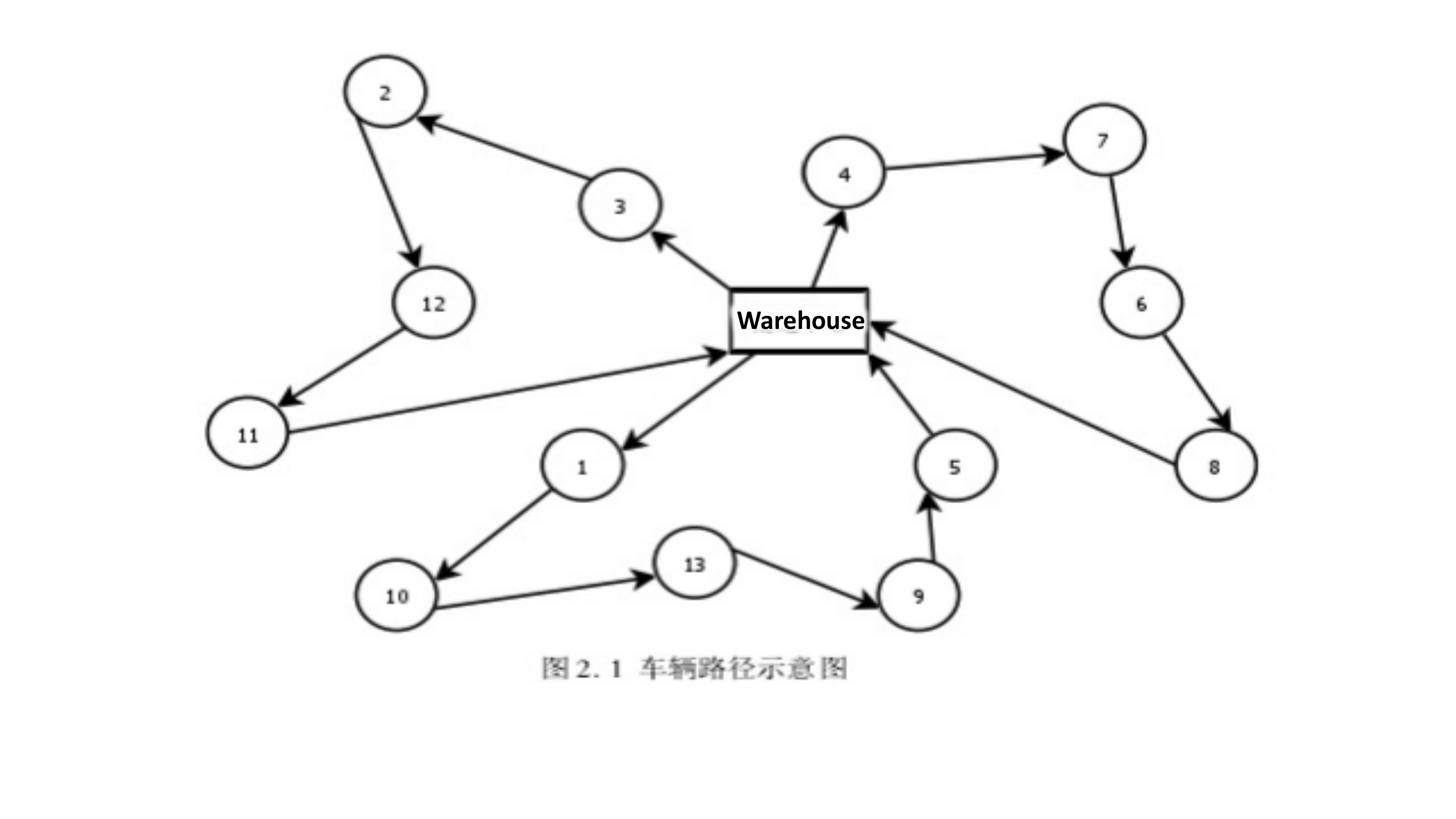}
  \caption{Routing planning problem sample.}
  \label{The_sample_of_Routing_planning_problem}
\end{figure}

The route planning problem refers that using the transportation equipment (in most cases, a car) to deliver goods from a fixed distribution centre to a dispersed number of customers and designing the most efficient transportation route to optimise some indicators, such as short delivery mileage, low total transportation cost, short transportation time, small number of usage of distribution vehicles, and high vehicle utilisation rate and so on. In this process, it needs to meet customers demand and other conditions. In general, the target is to minimise the distance and the number of vehicles. As shown in Figure~\ref{The_sample_of_Routing_planning_problem}, the three distribution devices start from the distribution centre, provide services to the customers in turn, and return to the distribution centre after completing all tasks.

Many researchers have explored a lot of route planning models and algorithms for this problem. Specifically, it can be divided into an exact algorithm and a heuristic algorithm. The exact algorithm can find the optimal solution. While heuristic algorithms generally give near-optimal answers to problems. Because computational complexity and computation time of all exact algorithms will increase exponentially with the increase of the problem size, the scope of use in the actual application process is very limited.  More research currently focuses on heuristic algorithms. All of the detail can be seen in the review of \cite{montoya2015literature}.

For the UAV's route planning, the optimisation of the path is similar to vehicle routing optimisation. So some basic algorithms are also used in this scenario. \cite{mittal2007three} propose the Evolutionary algorithms. In this method, not only the path length but also the maximum and minimum heights and collision risk were considerate. \cite{kennedy2010particle} introduce the Particle Swarm Optimisation method based on the scenario of reconnaissance. Several important factors that have an influence on the UAV route planning, such as safety, effectiveness, and target value are included in the objective function. The simulation can get satisfactory results. \cite{foo2009path} using Particle Swarm Optimization and B-splines to do a three-dimension path planning. The purpose of this planning is to minimise the risk of the enemy and minimise fuel consumption. Besides, the way to combine transport vehicles with drones has also attracted the attention of some scholars. \cite{ferrandez2016optimization} propose k-means and genetic methods to solve this kind of delivery problem. It has a contribution to saving in both energy and time.

In conclude,  scholars have done a  considerable amount of research on path planning methods, and many of the methods developed to solve vehicle routing problems are also used in the drones.  However, artificial intelligence has not been applied deeply. Therefore, this paper will use the method of reinforcement learning to study the path planning problem of drones.
\subsection{Artificial intelligence to solve route planning problem}

\cite{vinyals2015pointer} make the first attempt, who propose the pointer network. This model originated from the Sequence to Sequence model, which implements the function of converting one sequence into another, but does not require input sequences and output sequences to be equal in length, which means the number of outputs is variable and regardless of the input length. At the same time, added attention mechanism to this model, where the hidden state of the encoder is added to the hidden state of the decoder according to a certain weight. This improves the prediction accuracy of entire model. Researchers trained this model in a supervised manner and use it to output a sequence. It proved that can get a nearly optimal solution in the travel salesman problem.

Later, some scholars proposed another model to solve the combinatorial optimisation problem based on the pointer network. This model uses two recurrent neural networks for encoder and decoder. It describe a new calculation step called glimpses, which summarises the contributions of different parts of the input sequence. Using a supervised loss function to train this model. This framework has been applied to several classic combinatorial optimisation problems, such as travelling salesman problem, knapsack problem, and so on. In these experiments, this framework model has achieved good results \citep{bello2016neural}.

Recently, \cite{nazari2018reinforcement} propose an end to end framework to solve vehicle routing problem, which is a simplified version of Pointer network and based on the research of Vinyals et al. and Bello et al. The model uses a policy gradient algorithm to optimise network parameters by observing rewards. Finally, it can generate an optimal strategy. This method can obtain better quality results in a shorter time in solving the TSP problem. At the same time, in addressing the VRP problem, it is possible to get a better solution than the classical heuristic algorithm in the scenario where the node is less than 100. Our research is improved based on this approach.

Through the literature review above, in our research, We considered more efficient population distribution, and set up constraints which are more close to the UAV development, also considering the limitation of policy and natural, studying the route planning problem with UAV. The detail is discussed in the next section.

\section{Mathematical model}
\subsection{Hypotheses}
In this paper, We built a model based on reinforcement learning that handles the current problems of drones when delivering goods, including weight limits and detour problems caused by the no-fly zone. We have the following assumptions:

\begin{enumerate}
\item There is a distribution centre with a known location.

\item The customer location is known, the weight and volume of each customer’s goods is known.

\item The distribution centre has multiple drones, and the quantity can meet the demand.

\item The drone is a single model; the load capacity is the same and fixed.

\item The task of each customer can only be completed by one UAV, which means the demand does not allow to split.

\item The no-fly zone is round, having fixed area and no overlap.
\end{enumerate}
\subsection{Detour method}
\begin{figure}[!htb]
  \centering
    \includegraphics[width=6cm]{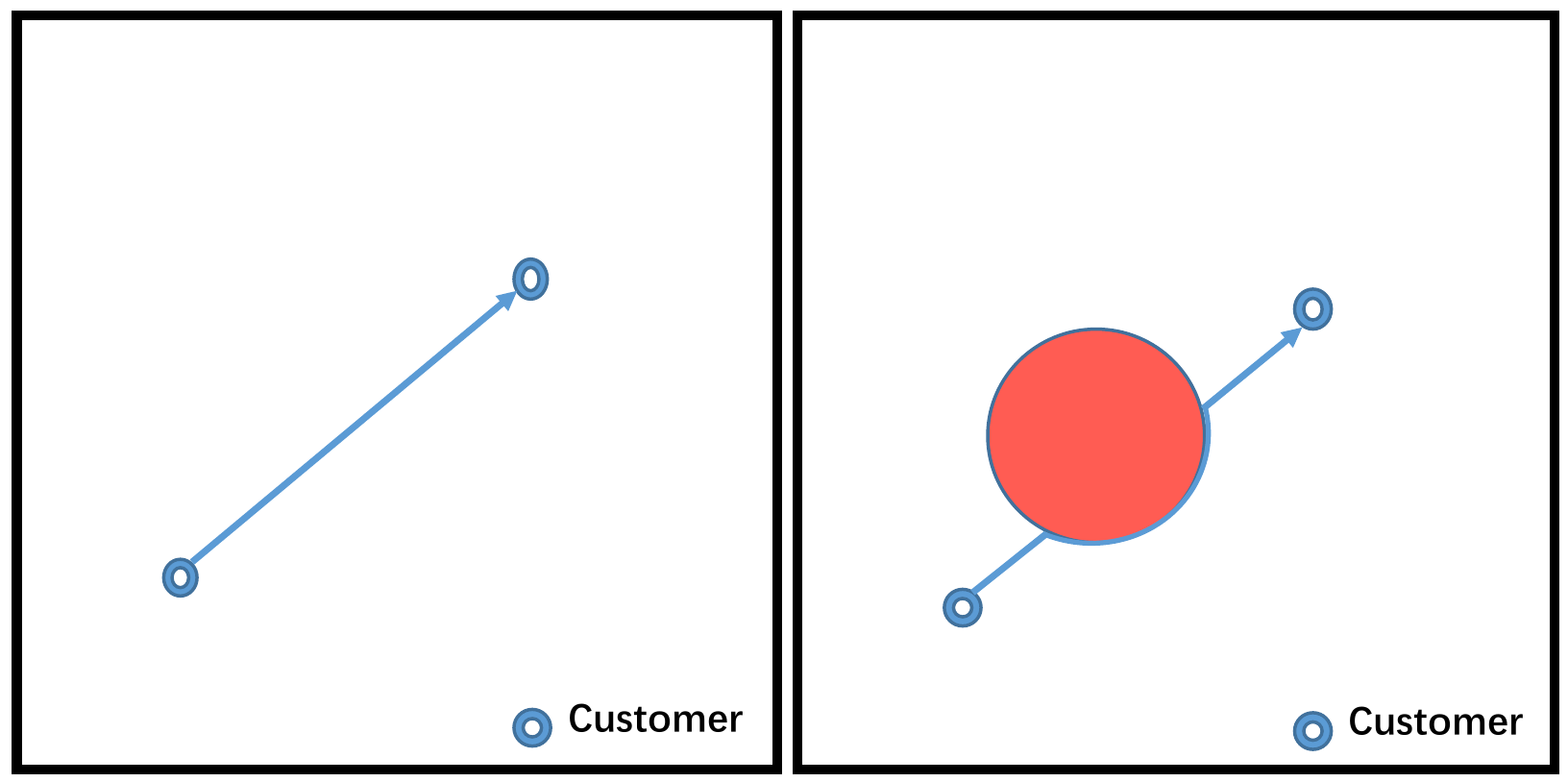}
  \caption{Detour Method (Red circle represents no-fly zone).}
  \label{detour_method}
\end{figure}

In general, the journey from one customer to another is a straight line. Because it has the shortest distance. However, when there are bad weather or regulations in certain areas of the air, the drone can only choose to bypass it to reach the destination. To address this problem, we set up a detour method, as shown in Figure~\ref{detour_method}, where red circle represents no-fly zone. When without the no-fly zone, the drone flies in the way of a straight line between two points. When UAV reaches the no-fly zone, it bypasses the shape of the no-fly zone.

The flight distance is calculated as follows.
\begin{equation}
    L=L_s+[\pi R/180^{\circ} \mathrm{min}(\theta _i,j,h ,(360^{\circ}-\theta _i,j,h))-l]
\end{equation}

In this equation, $L_s$ represents the straight distance of flying from one point to the next point. R is the no-fly circle's radius. $\theta _i,j,h$ represents the angle formed by the straight line between the two service points and the no-fly zone. $l$ represents UAV's flying distance in the no-fly region if we do not consider area control.

\subsection{Reinforcement learning model}

We solve this problem based on the model of Nazari's proposed in 2018. It has an excellent performance in the vehicle routing problem.

The purpose of this model is to find a strategy $\pi$that is infinitely close to the optimal strategy. The optimal strategy can generate an optimal solution in a probability of 100\%. We do not know what the optimal strategy is at first, so We often use a random strategy to experiment, and then we can get a trajectory consisting of a series of states, actions, and rewards. As follows:
\begin{equation}
    {s_1,a_1,r_1,s_2,a_2,r_2,...,s_t,a_t,r_t,s_{t+1}}
\end{equation}
Based on these trajectories, the strategy is improved continuously, and the rewards are continually increased. Finally, the performer can maximize the cumulative reward by performing actions according to this strategy.

In each round of operation, the environment needs to feedback a reward value to the Agent, enabling the network to adjust the parameters based on the reward value to obtain the maximum expectation of reward. The purpose of article is to get the shortest flight path, so awards are set to be the opposite of the total path length.

\subsubsection{Embedding}

The embedded layer can convert data into continuous low-dimensional vectors. Firstly, this input is encoded by an index. In this research, give each index a different coordinate and demand. Then, creating an embedded matrix that determines the length of the vector. This way of embedding is an affine transformation. The data is not necessarily in accordance with the original distance ratio to input, but do some processing and extract the useful characteristics of data in the high dimension. In most of the case, this method is computationally efficient, because the vector can be continuously updated during the training of deep neural networks. So the embedded layer is done to map our input data to the M-dimensional vector space.

\subsubsection{Attention mechanism}

The attention mechanism is similar to the human visual attention mechanism. By quickly scanning the image, human vision obtains the target area that needs to be focused on. Then we pay more attention to this area to get more detailed information about the target. In this mechanism, we use the embedding layer as the encoder and a recurrent neural network as the decoder. The variable-length vector $a_t$ is used in the model to extract information from the input data, which means the $a_t$ represents the correlation of each input point with the next decoding step. The influence of the location closer to decoding point is more significant than the influence of the location which is far away.

Assume that the input to the i state in the embedded layer is ${\overline{x}_t^i}=({\overline{s}^i},{\overline{d}_t^i})$, using $h_t$ to represent the memory state of the RNN at the decoding step t. The formula for $a_t$ is as follows.
\begin{multline}
    a_t=a_t({\overline{x}_t^i},h_t)=\mathrm{softmax}(u_t), \\
 where\quad u_t^i=v_a^T  \mathrm{tanh}(W_a[{\overline{x}_t^i};h_t])
\end{multline}
In this formula, represent the Two series of vector. $c_t$ describe as the conditional probabilities, which computed as
\begin{equation}
    c_t=\sum_{i=1}^M a_t^i{\overline{x}_t^i}
\end{equation}
Finally, through the softmax function, the result is mapped to the value of (0,1), which is the probability value of each point.
\begin{multline}
    P(y_{t+1}{\mid}Y_t,X_t)=\mathrm{softmax}({\Tilde{u}}_t^i), \\
where\quad {\Tilde{u}}_t^i=v_c^T \mathrm{softmax}(W_c[{\overline{x}_t^i};c_t])
\end{multline}
In these formulas, $v_a$, $v_c$, $W_a$, $W_c$ are variables that will change in the process of training.
\subsubsection{Decoder}
In the decoding stage, this model uses the strategy of beam search. This search method not only considers the highest probability point but also considers other multiple choices and has a parameter called beam width. This parameter represents the number of results retained in one calculation. For example, if the value equal to 3, the result of the top three probabilities will be retained after the search, and the result will be stored in the computer. The subsequent calculations will simultaneously calculate the probability of the result in three cases. Then we also pick up the first three probability result. Repeat this process until the end. In this algorithm, the value of beam width is bigger, the computational complexity of the algorithm is greater, but the quality of the solution may become better. 

\subsubsection{State transition}
A Markov decision process is used to represent the transition of the system state. The state of this phase is often the result of the previous phase state and the previous phase of the decision. We use the Yt to represent the decode sequence at time t, $Y_t$ = $y_0$, ..., $y_t$, While the $y_t$ describe the value of each decode time. The system update process is computed as
\begin{equation}
    X_t+1 = f(y_t+1,X_t)
\end{equation}
Specifically, assume that at the time point of 0, the drone is in the warehouse, and the first decoded point is the coordinates of the warehouse. This is the first state. Then, the drone selects the first customer and services it, and the system enters the second state. During this state transition, the demand for the customer point, the remaining load and volume of the drone change and need to be updated. The update function is as follows.
\begin{multline}
    d_{t+1}^i=\mathrm{max}(0,d_t^i-l_t),\quad d_{t+1}^i=d_t^i for t\not=i, \\
    and \quad l_{t+1}=\mathrm{max}(0, l_t - d_t^i)
\end{multline}

%

\subsection{Training}
This model is trained by a policy gradient method. It consists of two parts. The first is the actor network, which predicts the probability distribution of behaviour based on the function of softmax and selects the next action based on the distribution. The second is the critic network. It can evaluate the reward Q(s, a) value obtained in a state, which means in the state s, the expectation of the future reward will be obtained after taking action a.

\subsection{System operation process}
\begin{figure}[!htb]
  \centering
    \includegraphics[width=8cm]{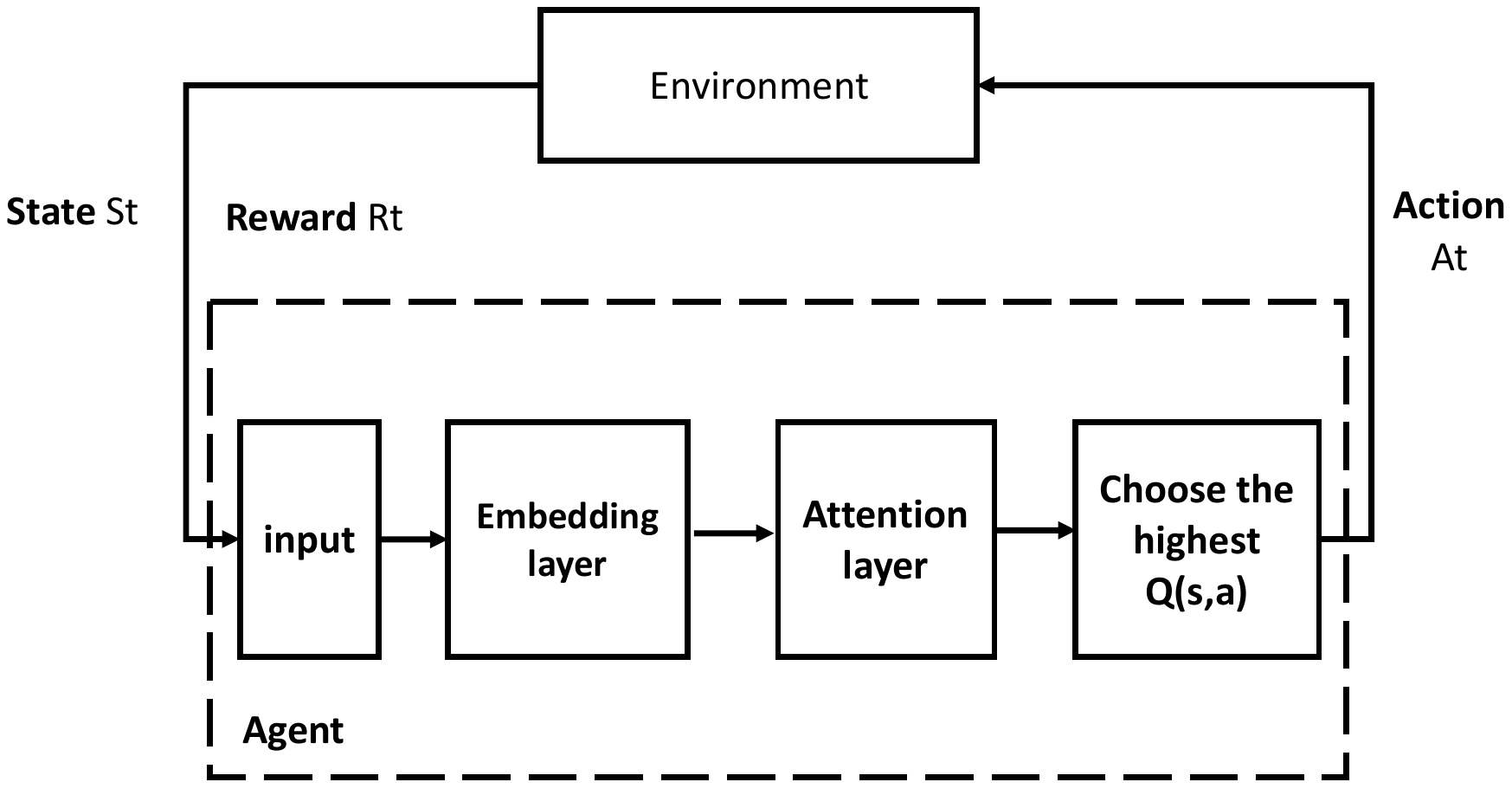}
  \caption{System operation process}
  \label{system_operation_process}
\end{figure}

As shown in the Figure~\ref{system_operation_process}, it is the system operation process diagram. In this research, the environment is the unmanned aerial vehicle. The agent is the part in the dotted box. The agent input data at the embedded layer and at the attention layer, calculate the possibility of selecting each feasible customer node as the next destination at that time point. Then select the customer point of the next service according to the corresponding decoding strategy. After acting on the environment, the system moves to the next state, and the environment feeds back to the Agent for a reward. The agent constantly adjusts its network parameters according to reward, making the calculated possibility of each customer more accurate. Finally, an optimal strategy is found, and the path is the shortest under the condition of completing the system constraint.

\section{COMPUTATIONAL EXPERIMENT}
\subsection{Data source}
At present, whether living in the city or countryside, the character of people living areas is that concentrated in a small area and a wide range of dispersion. So, to simulate the living conditions of people today, our research assumed that there are several places where people gather in a fixed-size area. We allocate 60\% of the population to the high-density population areas, and the remaining customers are evenly distributed anywhere in the fixed area. Through the python software to simulate the characteristics. Assign a different customer to produce probability to different regions. In the range of [1,1] generate distribution centres and demand points randomly. While, since the quality of each customer's purchased products cannot be predicted, we assume that the customer's needs or the weight of the delivered product meets a random distribution and the range is 0 to 9 units. We are using the random function in python to realize it. Besides, we fixed the coordinates of the warehouse to (0.5, 0.5) and the circle of the no-fly zone to (0.3,0.3). The area with a radius of 0.1.

This research simulates 1000 sets of three-dimension data sets as the training set and verification set of the model. The coordinate of 10, 20 and 50 customer node and demands in the inference set are shown in the Appendix.

\subsection{experimental setup}

The function of the model is realized using TensorFlow framework (Python 3.7). We improved the code that put forward by \cite{nazari2018reinforcement}. The main change of the system is the rules of data set generation, as well as the multiple round trips of a single device to the joint collaboration of multiple devices. Moreover, we set an area limitation that UAVs cannot fly. The relevant parameters in the experiment are set as follows.

We set the reward to the opposite of the total path length, because the purpose is to get the shortest flight path.

In beam search, the different beam widths represent the number of choices that are retained during each decoding process. The larger the number, the higher the quality of the calculated results. However, if the amount is too large, the calculation time will become too long. In this experiment, setting beam width to 1, 5 and 10, and then comparing the three cases.

In reinforcement learning, training sets and test sets are not distinguished. We completed training and testing the model at the same 1000 data sets. 

Moreover, the learning rate is set to $10^{-4}$. The batch size are set to 128 in this study. We set the dropout factor to 0.1, indicating that there is a 10\% probability that some options are not selected. 10, 20 and 50 are set to be the customer node respectively.

The training environment is in a central computer with 4 Intel(R) Xeon(R) E5-2620 CPUs (2.00GHz), an NVIDIA QUADRO RTX 6000 (24GB), and 24GB DRAM. With ten customers on a single GPU, each 200 training steps of a node's VRP takes approximately 35 seconds. The initial set of training steps is 50000, and the entire training time takes about 2.5 hours. The 20 customers model almost cost 100 seconds in each 200 training steps. The whole training time required about 6.9 hours. The 50 customers model needs 360 seconds to complete an update step. It cost 25 hours.

\subsection{Results}
After random testing, the results obtained are as follows Table~\ref{Table:The_Results} under various parameters.
\begin{table*}[!htb]
  \centering
  \begin{tabular}{clllllllll}   
  \toprule
  Baseline & 
  \multicolumn{3}{l}\textbf{C10,Cap30} & \multicolumn{3}{l}{C10,Cap40} & \multicolumn{3}{l}{C10,Cap50}
   \\ \cline{2-10} & mean(cm) & std(cm) & time(s) & mean(cm) & std(cm) & time(s) & mean(cm) & std(cm) & time(s) \\
 \midrule
  Beam Search(1) & 3.97 &0.63 &0.027 &3.58 & 0.57 & 0.018 & 3.35 & 0.52 & 0.017\\
  Beam Search(5) & 3.89 & 0.59 &0.029 & 3.49 & 0.54 & 0.020 & 3.28 & 0.49 &0.023 \\
  Beam Search(10) & 3.87 & 0.58 &0.035 & 3.45 & 0.53 & 0.026 & 3.24 & 0.47 & 0.027\\
 \bottomrule
 
 ~\\
   \toprule
  Baseline & 
  \multicolumn{3}{l}\textbf{C20,Cap30} & \multicolumn{3}{l}{C20,Cap40} & \multicolumn{3}{l}{C20,Cap50}
   \\ \cline{2-10} & mean(cm) & std(cm) & time(s) & mean(cm) & std(cm) & time(s) & mean(cm) & std(cm) & time(s) \\
 \midrule
  Beam Search(1) & 6.66 &0.87 &0.028 & 5.85 & 0.68 & 0.029 & 5.33 & 0.60 & 0.029\\
  Beam Search(5) & 6.52 & 0.85& 0.037 &5.68 & 0.65 & 0.039 & 5.21 & 0.57 & 0.043 \\
  Beam Search(10) & 6.42 &0.83 & 0.042  & 5.64& 0.66 &  0.056& 5.14 & 0.56 & 0.058\\
  \bottomrule
  
  ~\\
   \toprule
  Baseline & 
  \multicolumn{3}{l}\textbf{C50,Cap30} & \multicolumn{3}{l}{C50,Cap40} & \multicolumn{3}{l}{C50,Cap50}
   \\ \cline{2-10} & mean(cm) & std(cm) & time(s) & mean(cm) & std(cm) & time(s) & mean(cm) & std(cm) & time(s) \\
 \midrule
  Beam Search(1) &13.25 & 1.69 &0.078 & 11.89 & 1.52 & 0.072 & 10.79 & 1.46 & 0.071\\
  Beam Search(5) & 12.97& 1.65 &0.213 & 11.67 & 1.49 & 0.211 & 10.36 & 1.43 & 0.237\\
  Beam Search(10) & 12.83 & 1.63 &0.387 & 11.53 & 1.45 & 0.395 & 10.18 & 1.42& 0.401\\
  \bottomrule
  \end{tabular}
  \caption{The results of all experience.}
  \label{Table:The_Results}
\end{table*}

The inference data set are respectively brought into the corresponding trained model to calculate.  All of the model with a capacity of 30 and use a beam search method with a width of 10 to decode. The corresponding solution is shown in the Appendix.

\begin{figure}[!htb]
  \centering
    \includegraphics[width=6cm]{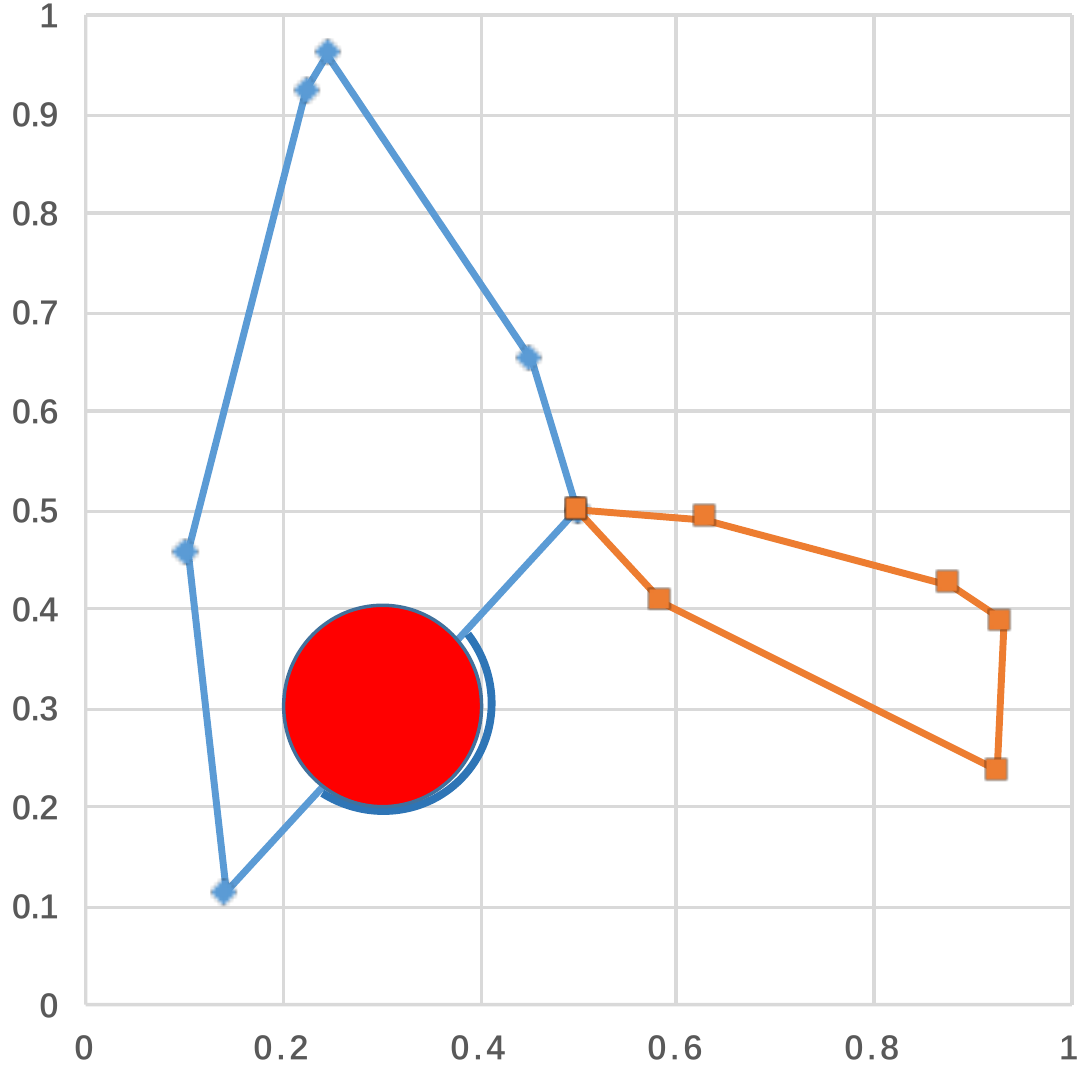}
  \caption{Visual sample solution of ten customers}
  \label{Figure:Visual_sample_solution_of_ten_customers}
\end{figure}

\begin{figure}[!htb]
  \centering
    \includegraphics[width=6cm]{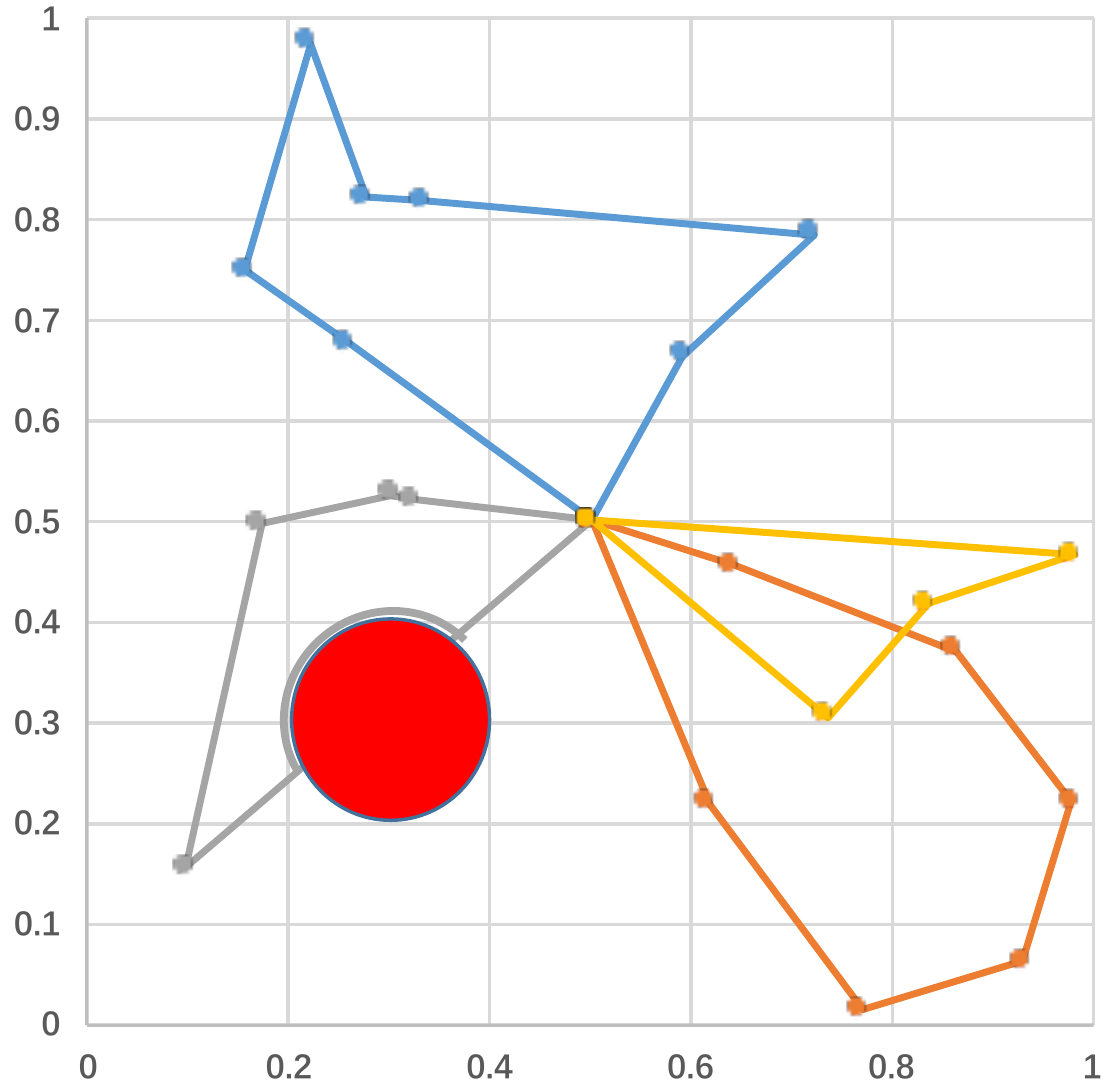}
  \caption{Visual sample solution of twenty customers}
  \label{Figure:Visual_sample_solution_of_twenty_customers}
\end{figure}

\begin{figure}[!htb]
  \centering
    \includegraphics[width=6cm]{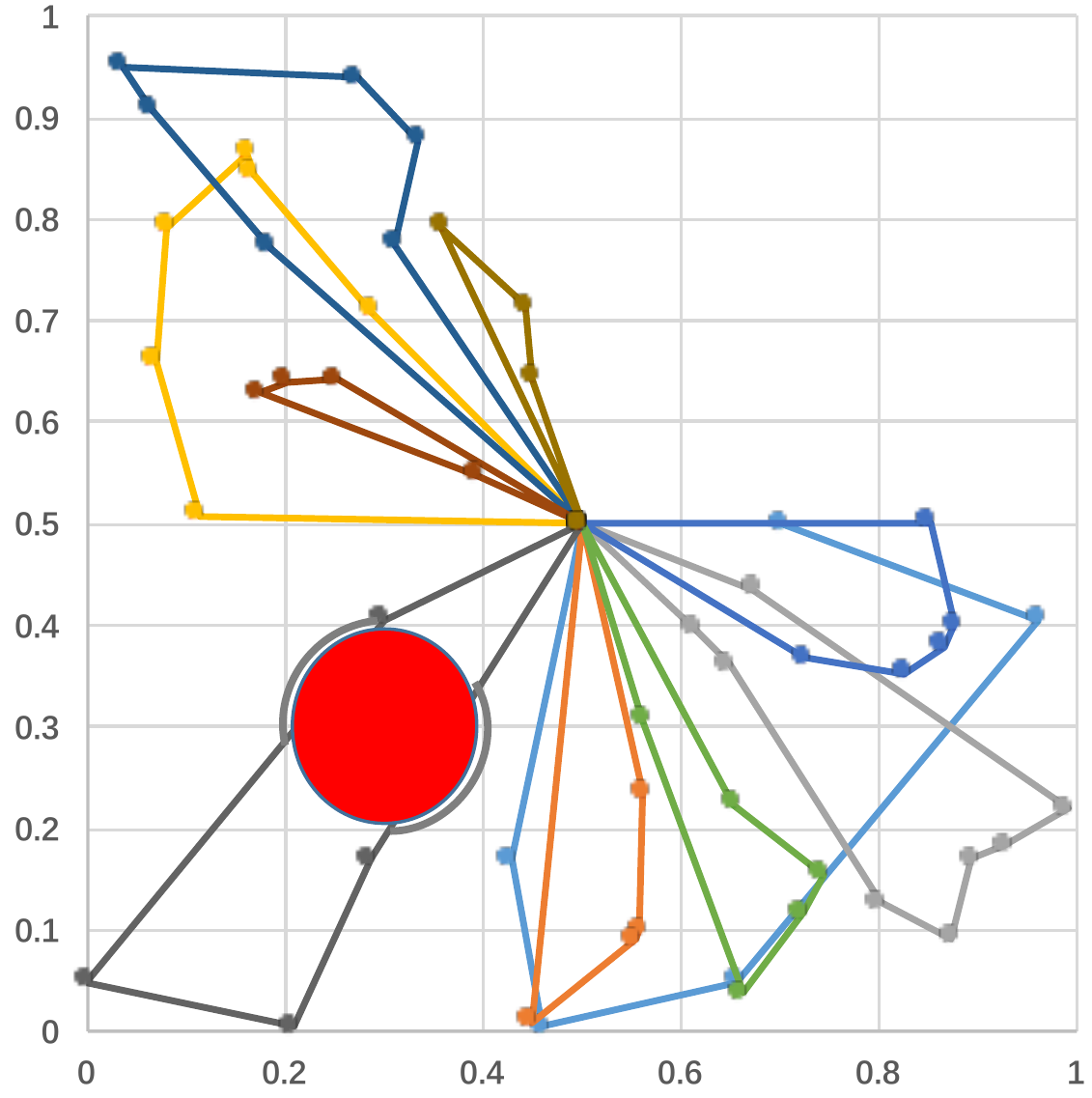}
  \caption{Visual sample solution of fifty customers}
  \label{Figure:Visual_sample_solution_of_fifty_customers}
\end{figure}

The route planning map is as shown in the Figure~\ref{Figure:Visual_sample_solution_of_ten_customers},Figure~\ref{Figure:Visual_sample_solution_of_twenty_customers} and Figure~\ref{Figure:Visual_sample_solution_of_fifty_customers}. The lines of different colors represent the flight routes of different drones, and the red circle represents no-fly zone.

\section{Discussion}
In the last section, the inference set is calculated in the model, and the validity is verified. Visually, under the constraints of the three different kinds of customer numbers, the drone completes the delivery task for each customer, which meet the system settings. And the weight of the goods carried by each drone in the plan is within the set feasible value. At the same time, when drones encounter a no-fly zone, it can choose detour routes. This aspect confirms the feasibility of this model. On the other hand, it also shows that the UAV distribution system can well meet the needs of customers and governments requirement and express the reliability of the distribution system.

\subsection{results comparison}

OR tools are Google’s open-source software for optimization issues. It can solve combinatorial optimization problems quickly and easily. And after building the model, we can use different kinds of language to solve it, including C++, Python, Java, and so on. It is currently one of the best solvers for solving vehicle routing problems. It uses a variety of heuristic algorithms to find a relatively optimal solution to the problem and avoid local optimal situations, such as the Newton method, gradient method, simulated annealing, genetic algorithm, ant colony algorithm and so on. It can represent a great solution to using the current heuristic algorithm to solve the route planning problem. So we plan to compare the result described in this paper with the result calculated by OR tools to test the quality of the solution.

To make the calculation result more accurate, We simulated ten sets of data and brought it into OR tools for calculation. The average value of the distance is calculated as a result. Since OR tools require the coordinates and requirements of the customer point can only be integers, so to solve this problem, we first enlarge the simulated number of customer coordinates by 10,000 times. Thus, the data is defined to be calculated in a square with 10,000 units of length and width. After the results are obtained, the path length is reduced by 10,000 times to compare with the method of this paper. The results shown in the Table~\ref{Table:OR_tools_solution} below.

\begin{table}[!htb]
  \centering
  \begin{tabular}{cccc}
  \toprule
  \textbf{Baseline} & \textbf{C10,Cap30}& \textbf{C10,Cap40}& \textbf{C10,Cap50}\\
  \midrule
  OR tools(cm) &3.62 &3.23 &2.98 \\
  \bottomrule
 
  ~\\
  \toprule
  \textbf{Baseline} & \textbf{C20,Cap30}& \textbf{C20,Cap40}& \textbf{C20,Cap50}\\
  \midrule
  OR tools(cm) & 6.29 &5.63 &5.09\\
  \bottomrule
  
  ~\\
  \toprule
  \textbf{Baseline} & \textbf{C50,Cap30}& \textbf{C50,Cap40}& \textbf{C50,Cap50}\\
  \midrule
  OR tools(cm) & 12.91 &11.38 &10.29 \\
  \bottomrule
  \end{tabular}
  \caption{The results of OR tools solution}
  \label{Table:OR_tools_solution}
\end{table}

Comparing the results in the Table~\ref{Table:OR_tools_solution} with the results in the Table~\ref{Table:The_Results}, we can find that when the number of customers node is small, such as the number of customers is 10 and 20 in this paper, the quality of the solution of OR tools is significantly higher than the quality of the model solution in this paper, but Within the standard deviation. When the number of customers node rises to 50, the solution quality of the OR-tools is degraded. The average distance becomes similar or larger than the distance calculated by the model described in this paper. The reason for this may be that the heuristic algorithm used by OR tools can use the traversal method to find the optimal solution of the problem when the number of points that need to be served is small, and it does not take too long. However, as the number of customers continues to increase, the amount that needs to be calculated rises in a straight line and will cost a lot of time. So it is no longer applicable, and to get a solution quickly, the quality of the solution has a little declined. While this paper uses the same model, so as the number of customers grows, the advantages gradually emerge.
Meanwhile, in the case of different capacities, the two solutions have the same trend, and the reasons are the same. Overall, the quality of the solution of this model is high, especially when the customer nodes are higher. So, it can meet path planning needs.
While, there is some disadvantage for the or-tools. Since the or-tools require us to input the distance matrix of each point, so we have to calculate customer's data by other methods before using. It is not convenient, especially for the no-fly zone and big data.

Besides, it can be seen that the beam search with width ten can get the highest quality of solutions. And the width one method gets the lowest quality of solutions in general in these three methods. This approves that the high number of width can effectively avoid local optimum situation and improves the quality of solutions.

\subsection{Algorithm performance comparison}
\subsubsection{Convergence analysis}

I set the system updated every two hundred steps and each model training 50000 steps. The calculation process of the algorithm in the different parameter shown in the Figure~\ref{Figure:Calculatin_process_of_different_parameter} below.
 
\begin{figure}[!htb]
  \centering
    \includegraphics[width=8cm]{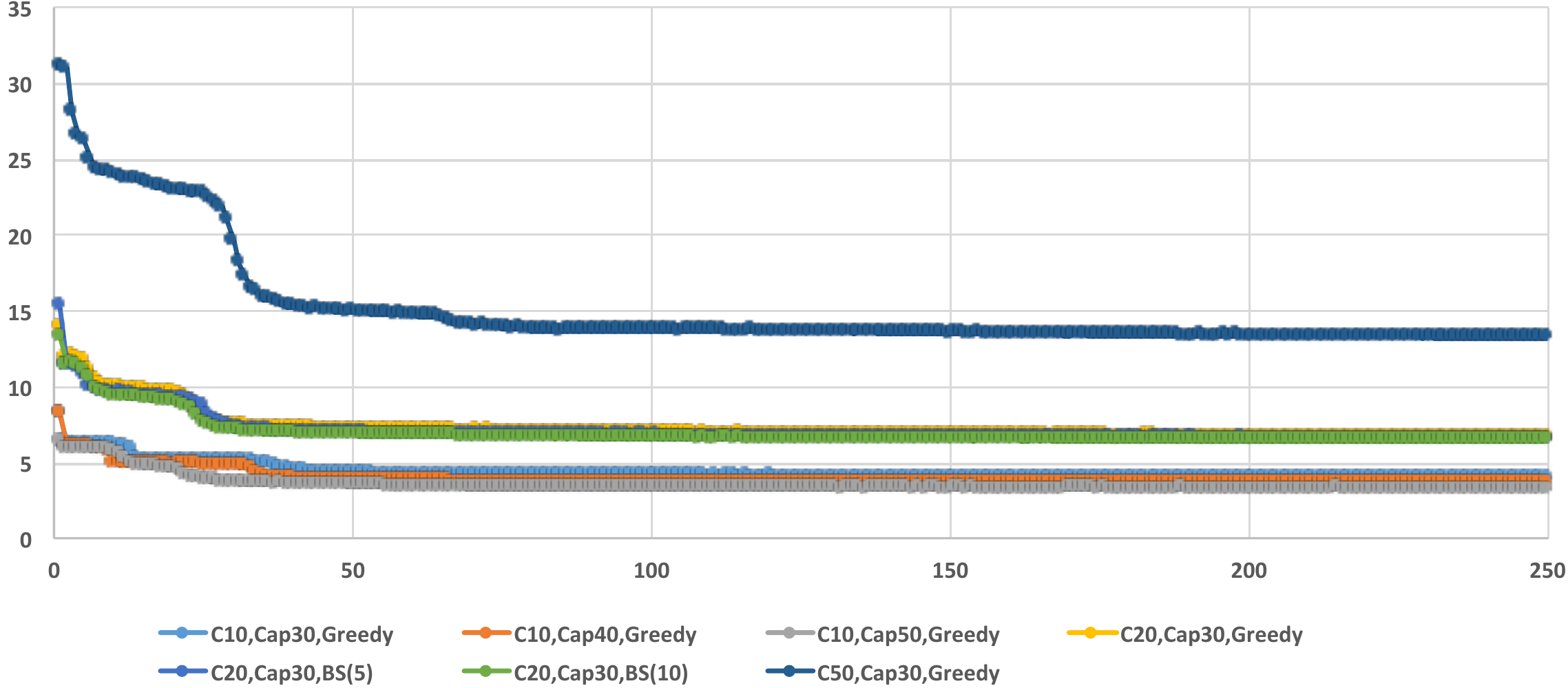}
  \caption{Calculation process of different parameter.}
  \label{Figure:Calculatin_process_of_different_parameter}
\end{figure}

As can be seen from the picture, all of the models almost have the same trend. In the beginning, it has a rapid decline. Then, there is a small drop with shock. Finally, the system reaches stable. Meanwhile, we also can see that there is a temporary stable in the first half of the picture. The reason is that it found an optimal local strategy at the current time. We compared the calculation process of different model and found that for different customer numbers, the number of steps required to find the optimal strategy increases as the number of customers increases. In the model with node 10, the relative optimal strategy has been obtained when iterating to 100 times. In a model with a node of 20, finding the optimal strategy requires 150 iterations. In a model with a node of 50, it needs to be iterated no less than 200 times. Besides, it can be seen from the model with 10 Customers and three different capacity limits that the change of capacity does not affect the speed of convergence. In the three situations, the model starts to converge at the same position. It also can be found that the decoding method does not influence the model convergence. This is proved by the 20 Customers’ model of beam search with width one, beam search with width five and beam search with width ten decoding methods in the picture. Overall, the model can get the optimal strategy after training with fewer steps, which is very efficient.

\subsubsection{Solution time analysis}
\begin{figure}[!htb]
  \centering
    \includegraphics[width=8cm]{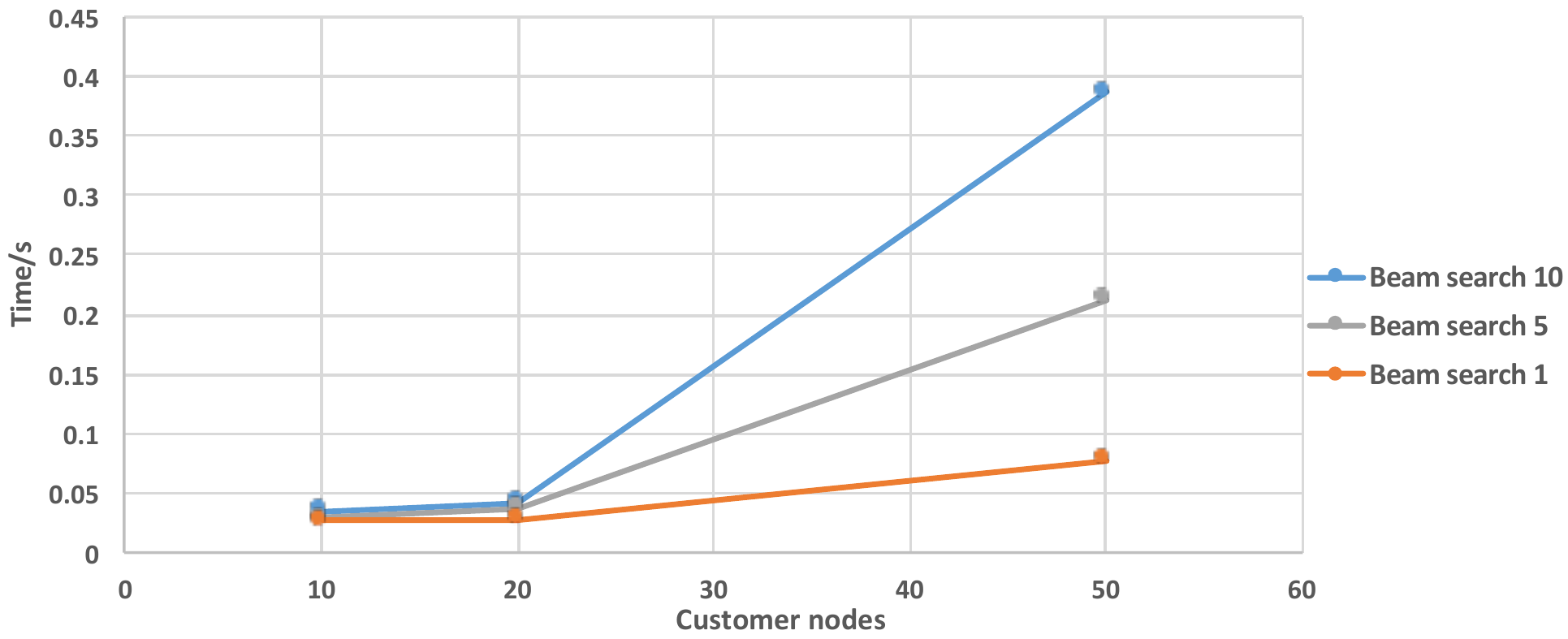}
  \caption{Node number and running time relationship.}
  \label{Figure:Node_number_and_running_time_relationship}
\end{figure}

The following analysis is about the effect of different parameters on the running time. According to the Table~\ref{Table:The_Results}, we draw a diagram about the relationship between the parameters and the running time as follows Figure~\ref{Figure:Node_number_and_running_time_relationship}.

The three different colour line in Figure~\ref{Figure:Node_number_and_running_time_relationship}  show the influence of the change in the number of customer nodes on the running time under the condition of varying decoding method when the capacity is 30. It can be seen that as the number of customer nodes increases, the calculation time is slowly increasing in the interval of 10 customers to 20 customers, but there is a considerable increase in the number of calculation time in the range of 20 customers to 50 customers. 

When the customer nodes are fixed, the calculation timeline increase with the rise of width for the beam search decoding method. The reason for the growth is that Beam search with width 10 needs to consider top 10 possibilities each time, while Beam search with width five only needs to find the top five possibilities. The more situations that need to be considered, the more paths can be generated, and the worse situation can be better avoided. So Beam search with width ten can produce the best quality solution in these three different decoding methods. But the running time is also longer. This time-growth trend is not obvious when the customer nodes less than 20. Because when the number of nodes is small, the number of possible combinations that can be generated is small, and the calculation time is not apparent. 

\subsubsection{Algorithm disadvantage}
If the model is applied in real life, there are still some disadvantages. On the one hand, the general performance of the model is not strong. It needs to use a different model to calculate the input of different parameters. In this paper, for each different number of customer nodes or different capacity limits, we need to train new models. However, in real life, the number of customers is constantly changing, and different types of drones also have different capacity ranges. This leads us to train a huge number of models to meet the needs to solve the whole question, also result in the high requirement of workload and the system storage capacity.
On the other hand, path security is another problem. The operation of the unmanned aerial vehicle mechanism is that after the set up of the system, the delivery process is only the execution of the planned path. The UAVs does not know the state of other drones in the air. Meanwhile, it also cannot deal with an emergency. As can be seen from the route diagram, there are many intersections in the planned paths, especially when the number of customer nodes is big. The crossed paths are likely to cause collisions of drones and create much potential safety risk. Therefore, the path planning for the drone should think more in terms of three-dimensional coordinates to minimize the overlap of the route and avoid the double loss of the aircraft and the customers' goods that result by the collision.

\section{Conclusion}

From the customer's point of view, we know that cost issue is an important reason why UAV has not been put into general operation. From the corporation's point of view, tighter policy constraints have also slowed its development. Finding a path planning method that can deal with those problems and planning a highly effective and proper route can significantly reduce the operating cost of the enterprise and increase the feasibility of using.
In conclusion, the proposed model and algorithm have good feasibility and stability, which has important guiding significance for the drones to avoid flying into no-fly zones and arrange a relatively shortest delivery route. The main contribution of this research is to add the consideration of residential distribution characteristics, strongly uncertain no-fly zones and multi-equipment coordination, which make the experiment more practical. In the future, we can concentrate more on three-dimensions research to improve the security of flying.

\appendices

\bibliographystyle{agsm}
\bibliography{access}

@article{zhang2012application,
  title={The application of small unmanned aerial systems for precision agriculture: a review},
  author={Zhang, Chunhua and Kovacs, John M},
  journal={Precision agriculture},
  volume={13},
  number={6},
  pages={693--712},
  year={2012},
  publisher={Springer}
}

@online{coombs2019amazon,
  author = {Casey Coombs},
  title = {Amazon’s Ambitious Drone Delivery Plans Take Shape},
  url = {https://www.thedailybeast.com/with-prime-air-amazon-wants-to-deliver-packages-in-30-minutes-or-less-via-drone}
}

@online{WIGGERS2019Wing,
  author = {KYLE WIGGERS},
  title = {Wing launches drone delivery in Christiansburg, Virginia},
  url = {https://venturebeat.com/2019/10/18/wing-launches-drone-delivery-in-christiansburg-virginia/},
}

@article{yu2018unmanned,
  title={Unmanned aerial vehicles: potential tools for use in zoonosis control},
  author={Yu, Qing and Liu, Hui and Xiao, Ning},
  journal={Infectious diseases of poverty},
  volume={7},
  number={1},
  pages={49},
  year={2018},
  publisher={BioMed Central}
}

@article{montoya2015literature,
  title={A literature review on the vehicle routing problem with multiple depots},
  author={Montoya-Torres, Jairo R and Franco, Juli{\'a}n L{\'o}pez and Isaza, Santiago Nieto and Jim{\'e}nez, Heriberto Felizzola and Herazo-Padilla, Nilson},
  journal={Computers \& Industrial Engineering},
  volume={79},
  pages={115--129},
  year={2015},
  publisher={Elsevier}
}

@inproceedings{vinyals2015pointer,
  title={Pointer networks},
  author={Vinyals, Oriol and Fortunato, Meire and Jaitly, Navdeep},
  booktitle={Advances in Neural Information Processing Systems},
  pages={2692--2700},
  year={2015}
}

@article{bello2016neural,
  title={Neural combinatorial optimization with reinforcement learning},
  author={Bello, Irwan and Pham, Hieu and Le, Quoc V and Norouzi, Mohammad and Bengio, Samy},
  journal={arXiv preprint arXiv:1611.09940},
  year={2016}
}

@inproceedings{nazari2018reinforcement,
  title={Reinforcement learning for solving the vehicle routing problem},
  author={Nazari, Mohammadreza and Oroojlooy, Afshin and Snyder, Lawrence and Tak{\'a}c, Martin},
  booktitle={Advances in Neural Information Processing Systems},
  pages={9839--9849},
  year={2018}
}

@inproceedings{mittal2007three,
  title={Three-dimensional offline path planning for UAVs using multiobjective evolutionary algorithms},
  author={Mittal, Shashi and Deb, Kalyanmoy},
  booktitle={2007 IEEE Congress on Evolutionary Computation},
  pages={3195--3202},
  year={2007},
  organization={IEEE}
}

@article{kennedy2010particle,
  title={Particle swarm optimization},
  author={Kennedy, James},
  journal={Encyclopedia of machine learning},
  pages={760--766},
  year={2010},
  publisher={Springer}
}

@article{foo2009path,
  title={Path planning of unmanned aerial vehicles using B-splines and particle swarm optimization},
  author={Foo, Jung Leng and Knutzon, Jared and Kalivarapu, Vijay and Oliver, James and Winer, Eliot},
  journal={Journal of aerospace computing, Information, and communication},
  volume={6},
  number={4},
  pages={271--290},
  year={2009}
}

@article{ferrandez2016optimization,
  title={Optimization of a truck-drone in tandem delivery network using k-means and genetic algorithm},
  author={Ferrandez, Sergio Mourelo and Harbison, Timothy and Weber, Troy and Sturges, Robert and Rich, Robert},
  journal={Journal of Industrial Engineering and Management (JIEM)},
  volume={9},
  number={2},
  pages={374--388},
  year={2016},
  publisher={Barcelona: OmniaScience}
}

\section{The coordinates of different number of customers}

\subsection{ten customers}

\begin{table}[!htb]
 \centering
 \setlength{\tabcolsep}{0.8mm}{
  \begin{tabular}{cccccccc}
  \toprule
  \textbf{No} & \textbf{Abscissa} & \textbf{Ordinate}  & \textbf{Demands}  & \textbf{No}  & \textbf{Abscissa} & \textbf{Ordinate}  & \textbf{Demands} \\
  \midrule
  1&    0.2455&    0.9603&    9&6&    0.4519&    0.6529&    1\\
2    &0.6302&    0.4914&    8&7    &0.1035&    0.4580&    5\\
3    &0.1411&    0.1143&    1 & 8    &0.9320&    0.3869&    9\\
4&    0.2235&    0.9241&    5 &9    &0.8790&    0.4236&    4\\
5    &0.5860&    0.4073&    6& 10    &0.9253&    0.2373&    3\\
  \bottomrule
  \end{tabular}}
  \caption{The coordinates of the ten customers}
  \label{Table:The coordinates of the ten customers}
\end{table}

\subsection{twenty customers}

\begin{table}[!htb]
 \centering
 \setlength{\tabcolsep}{0.8mm}{
  \begin{tabular}{cccccccc}
  \toprule
  \textbf{No} & \textbf{Abscissa} & \textbf{Ordinate}  & \textbf{Demands}  & \textbf{No}  & \textbf{Abscissa} & \textbf{Ordinate}  & \textbf{Demands} \\
  \midrule
 1&0.7225 &0.7845 &5&11&0.9310 &0.0610 &7\\
2&0.6434 &0.4553 &3&12&0.3023 &0.5254 &8\\
3&0.2593 &0.6761 &1&13&0.7356 &0.3057 &7\\
4&0.2214 &0.9741 &5&14&0.2761 &0.8218 &4\\
5&0.5923 &0.6647 &2&15&0.3231 &0.5213 &9\\
6&0.8368 &0.4175 &8 &16&0.9795 &0.2196 &6 \\
7&0.0986 &0.1547 &7 &17&0.1732 &0.4969 &5\\
8&0.6183 &0.2192 &3 &18&0.7710 &0.0123 &7\\
9&0.1567 &0.7493 &6 & 19&0.9816 &0.4650 &8 \\
10&0.3329 &0.8179 &4 & 20&0.8638 &0.3713 &3 \\
  \bottomrule
  \end{tabular}}
  \caption{The coordinates of the twenty customers}
  \label{Table:The coordinates of the twenty customers}
\end{table}

\subsection{fifty customers}

\begin{table}[!htb]
 \centering
 \setlength{\tabcolsep}{0.8mm}{
  \begin{tabular}{cccccccc}
  \toprule
  \textbf{No} & \textbf{Abscissa} & \textbf{Ordinate}  & \textbf{Demands}  & \textbf{No}  & \textbf{Abscissa} & \textbf{Ordinate}  & \textbf{Demands} \\
  \midrule
 1&0.0370 &0.9512 &8&26&0.5543 &0.0900 &7\\
2&0.0005 &0.0501 &8&27&0.5592 &0.1001 &6\\
3&0.0642 &0.9098 &5&28&0.4613 &0.9197 &7\\
4&0.0706 &0.6604 &6&29&0.5622 &0.2366 &3\\
5&0.0827 &0.7914 &2&30&0.5624 &0.3068 &6\\
6&0.1147 &0.5080 &7&31&0.6152 &0.3963 &1\\
7&0.1635 &0.8654 &4&32&0.6502 &0.3612 &3\\
8&0.1681 &0.8448 &8&33&0.6540 &0.2244 &1\\
9&0.1752 &0.6295 &7&34&0.6575 &0.9169 &9\\
10&0.1861 &0.7720 &2&35&0.6624 &0.0375 &8\\
11&0.2018 &0.6406 &7&36&0.6755 &0.4363 &2\\
12&0.2087 &0.0054 &4&37&0.7020 &0.7828 &5\\
13&0.2528 &0.6419 &9&38&0.7245 &0.1166 &5\\
14&0.2734 &0.9387 &2&39&0.7257 &0.3677 &3\\
15&0.2875 &0.1696 &7&40&0.7437 &0.1549 &9\\
16&0.2892 &0.7098 &2&41&0.8026 &0.1272 &8\\
17&0.3008 &0.4066 &9&42&0.8265 &0.3532 &4\\
18&0.3124 &0.7758 &5&43&0.8525 &0.5014 &5\\
19&0.3353 &0.8776 &6&44&0.8654 &0.3796 &7\\
20&0.3599 &0.7917 &7&45&0.8751 &0.0923 &1\\
21&0.3960 &0.5486 &1&46&0.8776 &0.4001 &7\\
22&0.4296 &0.9446 &3&47&0.8948 &0.1697 &5\\
23&0.4448 &0.7137 &8&48&0.9293 &0.1833 &4\\
24&0.4487 &0.0103 &4&49&0.9620 &0.9115 &5\\
25&0.4520 &0.6444 &9&50&0.9907 &0.2197 &6\\
  \bottomrule
  \end{tabular}}
  \caption{The coordinates of the fifty customers}
  \label{Table:The coordinates of the fifty customers}
\end{table}

\section{Sample solution}

\begin{table*}[!htb]
\centering
\setlength{\tabcolsep}{10mm}{
  \begin{tabular}{p{12cm}}
  \toprule
  \textbf{Sample solution for customer 10,capacity of 30} \\
  \midrule
  Tour lengths:3.29\\
  Best tour: UAV1:0 Load(0) $\xrightarrow{}$  2 Load(8) $\xrightarrow{}$  9 Load(12) $\xrightarrow{}$ 8 Load(21)$\xrightarrow{}$  10 Load(24) $\xrightarrow{}$  5 Load(30) $\xrightarrow{}$ 0 Load(30)\\UAV2:0 Load(0) $\xrightarrow{}$  3 Load(1) $\xrightarrow{}$  7 Load(6)$\xrightarrow{}$ 4 Load(11)$\xrightarrow{}$ 1 Load(20)$\xrightarrow{}$ 6 Load(21) $\xrightarrow{}$  0 Load(21)\\
  \bottomrule
  
  ~\\
  \toprule
  \textbf{Sample solution for customer 20,capacity of 30} \\
  \midrule
  Tour lengths:5.91\\
  Best tour: UAV1:0 Load(0) $\xrightarrow{}$  5 Load(2) $\xrightarrow{}$  1 Load(7) $\xrightarrow{}$10Load(11) $\xrightarrow{}$  14 Load(15) $\xrightarrow{}$  4 Load(20)$\xrightarrow{}$9 Load(26) $\xrightarrow{}$  3 Load(27) $\xrightarrow{}$  0 Load(27)\\ UAV2:0 Load(0) $\xrightarrow{}$ 8 Load(3) $\xrightarrow{}$  18 Load(10) $\xrightarrow{}$ 11 Load(17) $\xrightarrow{}$ 16 Load(23)$\xrightarrow{}$ 20 Load(26) $\xrightarrow{}$ 2 Load(29) $\xrightarrow{}$ 0 Load(29)\\ UAV3:0 Load(0) $\xrightarrow{}$  15 Load(9) $\xrightarrow{}$  12 Load(17) $\xrightarrow{}$ 17 Load(22)$\xrightarrow{}$ 7 Load(29)$\xrightarrow{}$  0 Load(29)\\UAV4:0 Load(0)$\xrightarrow{}$ 13 Load(7)$\xrightarrow{}$  6 Load(15) $\xrightarrow{}$ 19 Load(23)$\xrightarrow{}$ 0 Load(23)\\
  \bottomrule
  
  ~\\
   \toprule
  \textbf{Sample solution for customer 50,capacity of 30} \\
  \midrule
  Tour lengths:11.50\\
  Best tour: UAV1:0 Load(0) $\xrightarrow{}$ 22 Load(3)$\xrightarrow{}$ 28 Load(10)$\xrightarrow{}$  34 Load(19) $\xrightarrow{}$ 49 Load(24) $\xrightarrow{}$ 37 Load(29)$\xrightarrow{}$ 0 Load(29)\\UAV2:0 Load(0) $\xrightarrow{}$ 29 Load(3) $\xrightarrow{}$ 27 Load(9)$\xrightarrow{}$ 26 Load(16) $\xrightarrow{}$ 24 Load(20)$\xrightarrow{}$ 0 Load(20)\\UAV3:0 Load(0)$\xrightarrow{}$ 36 Load(2) $\xrightarrow{}$50 Load(8) $\xrightarrow{}$48 Load(12) $\xrightarrow{}$  47 Load(17) $\xrightarrow{}$ 45 Load(18)$\xrightarrow{}$ 41 Load(26) $\xrightarrow{}$ 32 Load(29)$\xrightarrow{}$31 Load(30) $\xrightarrow{}$  0 Load(30)\\UAV4:0 Load(0) $\xrightarrow{}$ 6 Load(7) $\xrightarrow{}$  4 Load(13) $\xrightarrow{}$ 5 Load(15)$\xrightarrow{}$ 7 Load(19)$\xrightarrow{}$ 8 Load(27) $\xrightarrow{}$  16 Load(29) $\xrightarrow{}$ 0 Load(29)\\UAV5:0 Load(0) $\xrightarrow{}$ 43 Load(5)$\xrightarrow{}$ 46 Load(12)$\xrightarrow{}$ 44 Load(19)$\xrightarrow{}$ 42 Load(23)$\xrightarrow{}$ 39 Load(26)$\xrightarrow{}$ 0 Load(26)\\UAV6:0 Load(0) $\xrightarrow{}$  10 Load(2)$\xrightarrow{}$  3 Load(7) $\xrightarrow{}$ 1 Load(15) $\xrightarrow{}$  14 Load(17)$\xrightarrow{}$ 19 Load(23)$\xrightarrow{}$  18 Load(28)$\xrightarrow{}$  0 Load(28)\\UAV7:0 Load(0) $\xrightarrow{}$13 Load(9) $\xrightarrow{}$ 11 Load(16) $\xrightarrow{}$  9 Load(23)$\xrightarrow{}$ 21 Load(24) $\xrightarrow{}$ 0 Load(24)\\UAV8:0 Load(0) $\xrightarrow{}$  15 Load(7) $\xrightarrow{}$ 12 Load(11) $\xrightarrow{}$ 2 Load(19)$\xrightarrow{}$  17 Load(28) $\xrightarrow{}$ 0 Load(28)\\UAV9:0 Load(0) $\xrightarrow{}$ 30 Load(6) $\xrightarrow{}$  35 Load(14) $\xrightarrow{}$ 38 Load(19) $\xrightarrow{}$  40 Load(28) $\xrightarrow{}$  33 Load(29)$\xrightarrow{}$  0 Load(29)\\UAV10:0 Load(0)$\xrightarrow{}$  25 Load(9) $\xrightarrow{}$  23 Load(17) $\xrightarrow{}$ 20 Load(24)$\xrightarrow{}$ 0 Load(24)\\
  \bottomrule
  \end{tabular}}
  \caption{Sample solution}
  \label{Table:Sample_solution}
\end{table*}

\EOD

\end{document}